\pdfoutput=1

\documentclass[11pt]{article}

\usepackage{authblk}
\usepackage{acl}

\usepackage{tikz}
\usepackage{pgfplots}
\usepackage{tabularx}
\usepackage{multirow}
\usepackage{booktabs}
\usepackage{enumitem}
\hypersetup{colorlinks, urlcolor={medium-blue}}

\definecolor{medium-blue}{rgb}{0,0,1}
\definecolor{rd}{HTML}{D7263D}
\definecolor{blu}{HTML}{255C99}
\definecolor{orng}{HTML}{E59500}

\usepackage{array}
\newcolumntype{H}{>{\setbox0=\hbox\bgroup}c<{\egroup}@{}}

\title{InPars: Data Augmentation for Information Retrieval \\using Large Language Models
}

\author[$\dagger\diamond\ddagger\star$]{\textbf{Luiz Bonifacio}}
\author[$\dagger\diamond\star$]{\textbf{Hugo Abonizio}}
\author[$\dagger\star$]{\textbf{Marzieh Fadaee}}
\author[$\dagger\diamond\ddagger\triangleleft\star$]{\textbf{Rodrigo Nogueira}}
\affil[$\dagger$]{Zeta Alpha}
\affil[$\diamond$]{NeuralMind}
\affil[$\ddagger$]{University of Campinas}
\affil[$\triangleleft$]{University of Waterloo}
\affil[$\star$]{\small All authors contributed equally.}

\begin{document}

\maketitle

\begin{abstract}
The information retrieval community has recently witnessed a revolution due to large pretrained transformer models. Another key ingredient for this revolution was the MS MARCO dataset, whose scale and diversity has enabled zero-shot transfer learning to various tasks.
However, not all IR tasks and domains can benefit from one single dataset equally. 
Extensive research in various NLP tasks has shown that using domain-specific training data, as opposed to a general-purpose one, improves the performance of neural models~\cite{yu2021adaptsum,sharami2022selecting}.
In this work, we harness the few-shot capabilities of large pretrained language models as synthetic data generators for IR tasks.
We show that models finetuned solely on our unsupervised dataset outperform strong baselines such as BM25 as well as recently proposed self-supervised dense retrieval methods.
Furthermore, retrievers finetuned on both supervised and our synthetic data achieve better zero-shot transfer than models finetuned only on supervised data. 
Code, models, and data are available at \url{https://github.com/zetaalphavector/inpars}
\end{abstract}

\section{Introduction}

Language models (LMs) such as GPT-3~\cite{DBLP:journals/corr/abs-2005-14165}, FLAN~\cite{wei2022finetuned}, Gopher~\cite{rae2021scaling}, and T0++~\cite{sanh2021multitask} have demonstrated impressive performance on many NLP tasks.
Additionally, when sufficient supervised information is not available for a task, they have been shown to be effective and at times yield compelling results~\cite{winata-etal-2021-language,schick-schutze-2021-just}.

Despite the appealing capabilities of large LMs, multi-billion parameter models are rarely used in information retrieval (IR), with some notable exceptions~\cite{nogueira2020document,pradeep2021expando,neelakantan2022text}.
One reason is the computationally intensive nature of information retrieval tasks. In a typical reranking task, for instance, we compute the relevancy of 1000 candidate documents for one query, which requires 1000 inference passes on a reranking model. This can be prohibitively expensive when using large models. For example, OpenAI offers a search API that allows one to compute query-document relevancy using their models with billions of parameters. As of February 2022, they charge 0.06 USD per 1000 tokens for their largest model.
%
%
\begin{figure}
  \includegraphics[width=0.45\textwidth]{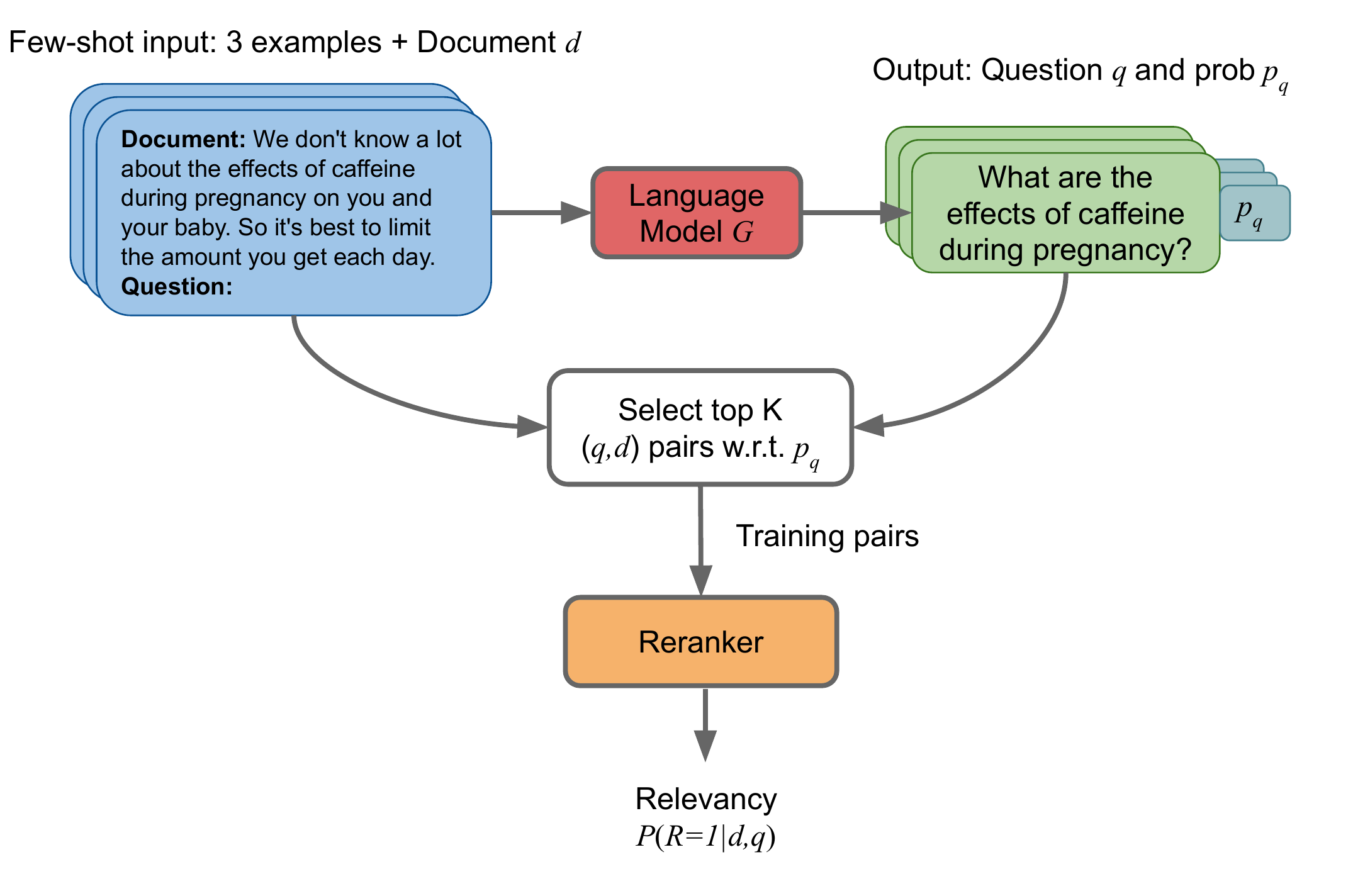}
  \caption{Illustration of our few-shot method that generates training data for search tasks. We use a language model $G$ to generate a question $q$ (and its probability $p_q$) from a document $d$. The top $K$ pairs $(q,d)$ with respect to $p_q$ are used as positive examples to train a reranker whose task it to estimate the relevancy of $d$ to $q$.}
  \label{fig:overview}
\end{figure}
If each candidate document contains 250 tokens, naively using this API for a reranking task would cost approximately 15 USD per query.

Dense retrievers~\cite{DBLP:journals/corr/abs-2004-04906, DBLP:journals/corr/abs-2004-12832} avoid this expensive reranking step by precomputing vector representations for each document in the collection prior to retrieval. When a query comes in, only its vector representations are computed, and a fast vector search framework can be used to retrieve the nearest document vectors to the vector representation of the query~\cite{johnson2019billion}. Despite being computationally cheaper at inference time, dense retrievers need one inference pass to compute the vector representation of each document in the collection, which also makes billion-parameter neural models impracticable to be used as dense retrievers.\footnote{For a more detailed cost estimation, check this \href{https://medium.com/@nils_reimers/openai-gpt-3-text-embeddings-really-a-new-state-of-the-art-in-dense-text-embeddings-6571fe3ec9d9}{blog post}.}

Another challenge in developing neural models for IR is the lack of domain-specific training data. Manually constructing high-quality datasets is difficult as it requires queries from real users. 
While there are a few general-purpose labeled data available~\cite{DBLP:journals/corr/NguyenRSGTMD16,47761}, they are not always effective in generalizing to out-of-domain datasets~\cite{thakur2021beir}.
For this goal, zero-shot and few-shot learning models are in particular promising.
However, a cost-effective manner of using large LMs in IR tasks is still an open question.

In this work, we propose a simple yet effective approach towards efficiently using large LMs in retrieval and obtain improvements across several IR datasets.
Rather than using large LMs directly in the retrieval process, we harness them to generate labeled data in a few-shot manner.
We then finetune retrieval models on this synthetic data and use them to rerank the the search results of a first-stage retrieval system. 
\setitemize{noitemsep,topsep=0pt,parsep=0pt,partopsep=0pt}
We summarize our contributions as follows:
\begin{itemize}
    \item We propose a method for adapting large LMs to IR tasks that otherwise are infeasible to be used due to their computational demands.
    \item In an unsupervised setting, our method largely outperforms recently proposed ones. When combined with supervised finetuning, our method achieves state-of-the-art results in two of the three transfer learning datasets evaluated in this work.
\end{itemize}

\section{Related Work}

Data augmentation methods aim at increasing the amount of data to assist the learning process of data-driven models.
To improve the performance of neural models in low-resource settings, small-scale LMs have been used to generate synthetic data in various NLP tasks~\cite{fadaee-etal-2017-data,kobayashi2018contextual}. 
Recent works show that large pretrained LMs are capable of generating data of reasonable quality~\cite{anaby2020not,papanikolaou2020dare,yang2020g,mohapatra2021simulated,kumar2020data,schick2021generating,meng2022generating}, sometimes leading to better transfer learning than human generated datasets~\cite{liu2022wanli}.

In information retrieval, dense retrievers can achieve comparable results to BM25 in some datasets when solely pretrained on documents without annotations~\cite{ram2021learning,izacard2021towards,neelakantan2022text}. These methods rely on extracting pairs of segments of texts that are likely relevant to each other which then used as positive pairs to train the retrieval models.

Focusing on improving the transfer learning effectiveness of dense retrievers, \citet{ma-etal-2021-zero} and \citet{wang2021gpl} use supervised sequence-to-sequence models to augment the training data. They generate questions from texts from different collections and use these synthetic question-text pairs as positive training examples.
Our work differs from existing approaches as we rely exclusively on simple prompts to generate questions from large language models with minimal supervision, i.e., using only a few supervised examples.
We were mostly inspired by~\citet{han2021unsupervised}, who uses such models to generate synthetic translation pairs in a zero-shot manner, i.e., without using any parallel corpora.

\section{Our Method: \textit{InPars}}
\label{sec:method}

In this section, we describe the proposed method, dubbed \textit{InPars} (Inquisitive Parrots for Search), for generating synthetic training datasets for IR tasks.
Given a document $d$ and a prefix $t$ consisting of $N$ pairs of questions and their relevant documents, i.e., $t=\{(q_1^*,d_1^*), ..., (q_N^*,d_N^*)\}$, our method uses a language model $G(t,d)$ to generate a question $q$ that is likely to be relevant to $d$.
The pair $(q, d)$ forms a positive training example that is later used to finetune our retrieval models.

We generate thousands of these positive training examples using documents randomly sampled from a collection $D$. The prefix $t$ is always the same regardless of the input document $d$, i.e., we can potentially generate millions of synthetic training examples using only $N$ manually annotated examples. This characterizes our method as a few-shot learning approach as long as N is small (in our experiments, we use three examples).

As a last step to create our training dataset, we select the top $K$ pairs with respect to the following (log) probability:
\begin{equation}
    p_{q} = \frac{1}{|q|} \sum_{i=1}^{|q|} \log p(q_i|t,d,q_{<i}),
\end{equation}
\noindent where $p(q_i|t,d,q_{<i})$ is the probability assigned by $G$ when autoregressively generating the $i$-th token of $q$.
We show in Section~\ref{sec:filtering_ablation} that this filtering step largely improves IR metrics. 

Due to its few-shot nature, our method can be used to adapt retrievers to any collection or IR task, which we later empirically confirm on various collections. This is particularly relevant for IR tasks as gathering data to train retrieval models is an expensive process~\cite{yilmaz2020reliability}, with most high-quality collections having less than a few hundred queries~\cite{Voorhees1999OverviewOT,Voorhees2004OverviewOT}.

We do not perform any pretraining to adapt the model to the target corpus, such as proposed by \citet{gao2021unsupervised}. Our method does not require any modifications in the loss function, as is done by~\citet{izacard2021towards,neelakantan2022text}. This makes InPars also suitable for non-neural retrieval algorithms.


\section{Experimental Setup}

In this section, we describe the datasets used in this work, the procedure to generate questions from the datasets in an few-shot manner, and finally, how we train retrievers on this synthetic data.

\subsection{Datasets}

\textbf{MS MARCO} \citet{DBLP:journals/corr/NguyenRSGTMD16} is a large-scale ranking dataset with more than half million anonymized questions sampled from Bing's search query logs. Its passage ranking version is formed by 8.8M passages and approximately 500k training pairs of queries and relevant documents. On average, there is one relevant passage per query, which were manually annotated. The development and test sets contain approximately 6,900 queries each and the test set is kept hidden from the public.

\textbf{TREC-DL} \citet{DBLP:journals/corr/abs-2102-07662} is a dataset that uses the same collection of passages from MS MARCO, but it contains only 54 queries and a higher number of judged documents per query.

\textbf{Robust04} \cite{96071} is an retrieval dataset formed by 249 queries. Its corpus consists of 528k documents from the news wire domain. It has on average 1250.6 judged documents per query. 

\textbf{Natural Questions} \cite{47761} is an open domain Question Answering dataset created from Wikipedia documents. The questions are real anonymized queries submitted to Google's search engine. In this work, we use the BEIR's version of NQ, which contains 3,452 queries and 1.21 relevant documents per query in its development set. The corpus consists of 2.6M passages from Wikipedia.

\textbf{TREC-COVID} \cite{10.1093/jamia/ocaa091} dataset was created during a challenge organized by NIST whose goal was to study information retrieval methods in response to the pandemic. We use the dataset version provided by the BEIR benchmark. The dataset consists of 50 queries, an average of 1,326 judged documents per query, and a corpus of 171k articles from the COVID-19 scientific literature. 


\begin{figure*}
\begin{center}
    \includegraphics[width=\textwidth]{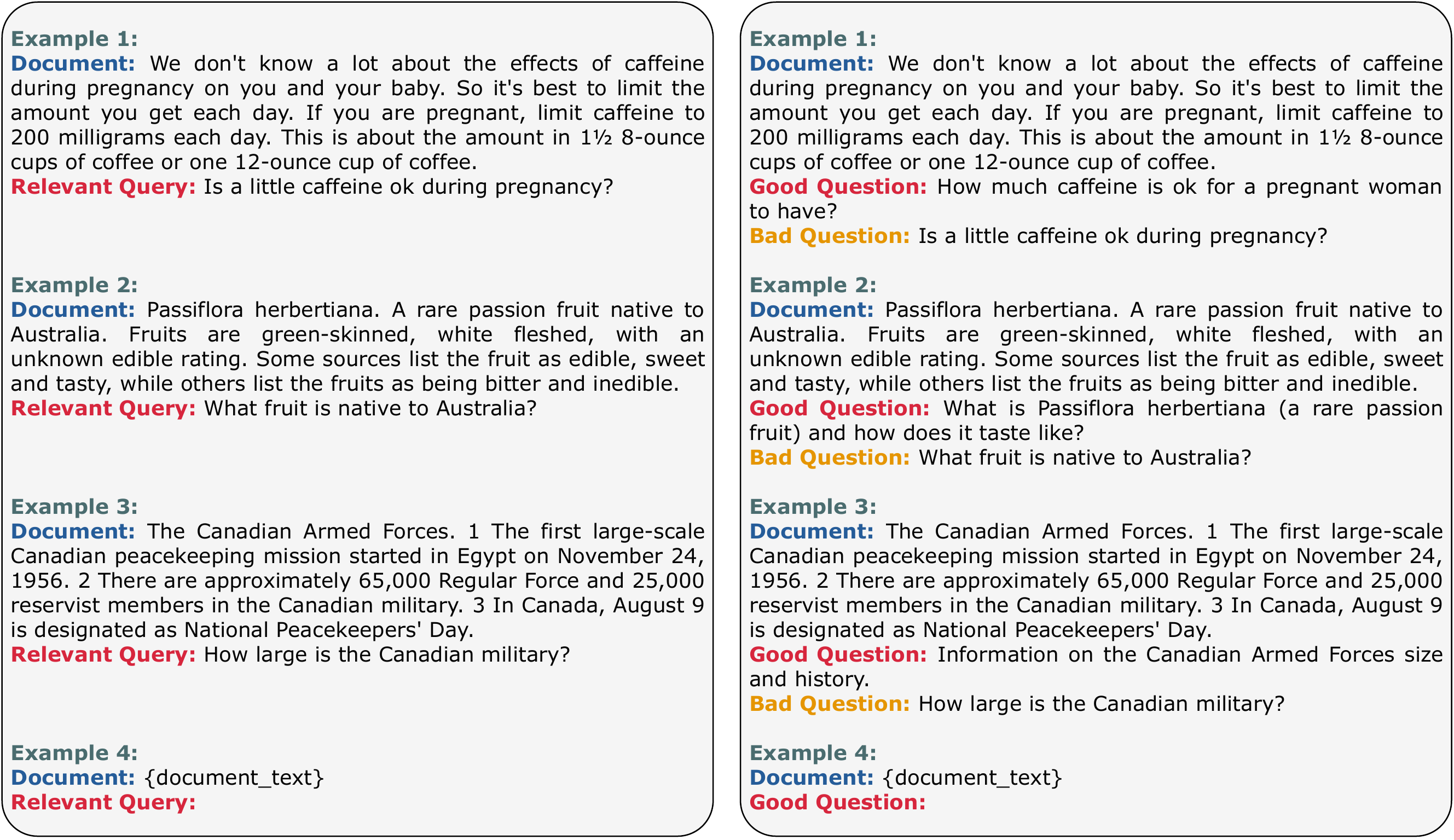}
  \caption{``Vanilla'' (left) and ``GBQ'' (right) prompts proposed in this work. The GBQ prompt consists of 3 relevant passages and queries randomly sampled from MS MARCO. The query is used as a \textit{bad question} and we provide a more descriptive \textit{good question} for the passage. 
  For the fourth example, we replace \texttt{\{document\_text\}} with a sampled document for which the language model is asked to generate a \textit{good question}.}
  \label{fig:prompts}
  \end{center}
\end{figure*}

\subsection{Training Data Generation}

Our training set comprises triples of a query, a positive, and a negative document. We first describe how pairs of query and positive documents are generated.
We randomly sample 100,000 documents from the collection and generate one question per document using GPT-3's Curie as our language model $G$. We prepend the document text with its title, when it is available. Documents with less than 300 characters are discarded and a new one is sampled.
We use a temperature of 0, which defaults to greedy decoding. We tried other decoding algorithms such as sampling (with a temperature of 1.0) and top-p sampling~\cite{holtzman2019curious} (with p=0.95) but did not notice significant differences in the final results.

We experiment with two prompt templates for generating questions which are illustrated in Figure~\ref{fig:prompts}.
The first called ``Vanilla'' (left), uses $N=3$ pairs of document and relevant question randomly chosen from the MS MARCO training dataset. The string \texttt{\{document\_text\}} is replaced with the document sampled from the collection and the language model generates a question one token at a time until a termination token \texttt{\textbackslash n} is chosen or a maximum number of 64 tokens is reached.

The second prompt template called ``Guided by Bad Questions'' (GBQ) illustrated in Figure~\ref{fig:prompts} (right) is similar to the ``Vanilla'' template, but we encourage the language model to produce more contextual-aware questions than the ones from MS MARCO. For that, we use MS MARCO questions as examples of ``bad'' questions, and manually create more complicated ones as examples of ``good'' questions. Rather than finding the answer in \textit{part} of the input document, the full context of the document will contribute to the answer.
With this prompt, the model generates a ``good'' and a ``bad'' question for each document and we keep only the good ones to form our positive training pairs.
Because the GBQ prompt, by design, generates questions that are different from MS MARCO ones, we use the ``Vanilla'' prompt to generate questions that are used in finetuning retrievers evaluated on MS MARCO and TREC-DL datasets.

There are four GPT-3 models available to the public via the OpenAI API: Ada, Babbage, Curie and Davinci. Their sizes are not explicitly mentioned, but from the figures reported by \citet{neelakantan2022text} and its accompanying blog post,\footnote{\url{https://openai.com/blog/introducing-text-and-code-embeddings/}} we infer that Ada, Baggage, Curie and Davinci have 300M, 1.2B, 6B and 175B parameters, respectively.
We show in Section~\ref{sec:model_size} that questions generated by the more expensive Davinci model lead to marginal improvements in IR metrics. Therefore, throughout this paper, we report results using questions generated by the Curie model.

Of the 100,000 questions generated (and their respective input documents), we use the top $K$=10,000 pairs w.r.t to $p_q$ as positive examples for finetuning our models.

Our retrievers are binary classifiers, so we also need to select documents that are not relevant to $q$ to form negative finetuning examples $(q, d^-)$.
We use a simple method that has been shown to be effective for finetuning rerankers~\cite{pradeep2021expando}.
We use BM25 with $q$ as query to retrieve 1000 documents from the collection $D$. We randomly select one of these as $d^-$, and the pair $(q,d^-)$ forms a negative example.

\subsection{Retrieval Methods}

We use a multi-stage retrieval architecture~\cite{matveeva2006high,wang2011cascade,chen2017efficient,liu2017cascade} comprised of initial retrieval with bag-of-words BM25~\cite{robertson1995okapi} followed by a neural reranker.

The collection is indexed using pyserini~\cite{lin2021pyserini} and 1000 candidate documents for each query are retrieved using BM25. 
Then we rerank the candidate documents using monoT5, which is an adaption of the T5 model~\cite{raffel2020exploring} to text ranking proposed by~\citet{nogueira2020document}.
We finetune monoT5 base (220M parameters) and 3B with a constant learning rate of $10^{-3}$ and an equal number of positive and negative examples in each batch of size 128. We did not conduct experiments with the 11B version due to its computational cost. To simplify our training procedure (and related hyperparameters) as well as to eliminate the need for convergence checks, we simply train for 156 steps (approximately one epoch).
Because the number of training steps is small, we observed a difference of up to two points in the final metric depending on the order in which the training examples are shown to the model. Thus, we report the results that are the average of three seeds.
Training T5 base and 3B takes a few minutes on a single Google TPU v3-8.

We finetune a model per collection using the synthetic questions generated from that collection. 
In Section~\ref{sec:prompt_ablation}, we show that having in-domain synthetic sets is beneficial. 
Nevertheless, using only MS MARCO as a source for documents also leads to reasonable results.
 
We use a maximum of 512 input tokens and two output tokens (one for the target token and another for the end-of-sequence token). In the experiments with the MS MARCO passage dataset, none of the inputs exceed this length limitation. For NQ and TREC-COVID, documents are often truncated. 

\begin{table*}
\centering\centering\resizebox{1.0\textwidth}{!}{
\begin{tabular}{llrrrrrrrH}
\toprule
& &\multicolumn{1}{c}{\textbf{MARCO}} & \multicolumn{2}{c}{\textbf{TREC-DL 2020}} & \multicolumn{2}{c}{\textbf{Robust04}} & \multicolumn{1}{c}{\textbf{NQ}} & \multicolumn{1}{c}{\textbf{TRECC}} & \multicolumn{1}{H}{\textbf{FiQA}}\\
& & \textbf{MRR@10} & \textbf{MAP}    & \small\textbf{nDCG@10} &  \textbf{MAP} & \small\textbf{nDCG@20} & \small\textbf{nDCG@10} & \small\textbf{nDCG@10} & \small\textbf{nDCG@10}\\
\midrule
\midrule
& \multicolumn{3}{l}{\textit{\textbf{Unsupervised}}} \\
(1) & BM25        & 0.1874 & 0.2876  & 0.4876 & 0.2531 & 0.4240 & 0.3290 & 0.6880 & 0.2536 \\
(2) & Contriever~\cite{izacard2021towards}  & - & - & - & - & - & 0.2580 & 0.2740 & 0.2450 \\
(3) & cpt-text~\cite{neelakantan2022text}  & 0.2270 & - & - & - & - & - & 0.4270 & 0.3970 \\
\midrule
& \multicolumn{4}{l}{\textit{OpenAI Search reranking 100 docs from BM25}}\\
(4) & Ada (300M)    & \$ & 0.3141 & 0.5161 & 0.2691 & 0.4847 & 0.4092 & 0.6757 & 0.3388\\
(5) & Curie (6B)    & \$ & 0.3296 & 0.5422 & 0.2785 & 0.5053 & 0.4171 & 0.7251 & 0.4012 \\
(6) & Davinci (175B) & \$ & 0.3163 & 0.5366 & 0.2790 & 0.5103 & \$ & 0.6918 & \$ \\
\midrule
& \textit{InPars (ours)}\\
(7) & monoT5-220M & 0.2585 & 0.3599 & 0.5764 & 0.2490 & 0.4268 & 0.3354 & 0.6666 & 0.3316\\ 
(8) & monoT5-3B & \textbf{0.2967} & \textbf{0.4334} & \textbf{0.6612} & \textbf{0.3180} & \textbf{0.5181} & \textbf{0.5133} & \textbf{0.7835} & \textbf{0.4103} \\
\midrule
\midrule
& \multicolumn{3}{l}{\textbf{\textit{Supervised} [$\triangleright$ MARCO]}} \\
(9) & Contriever~\cite{izacard2021towards} & - & - & - & - & - & 0.4980 & 0.5960 & 0.3290 \\
(10) & cpt-text~\cite{neelakantan2022text} & - & - & - & - & - & - & 0.6490 & 0.5120 \\
(11) & ColBERT-v2~\cite{santhanam2021colbertv2} & 0.3970 & - & - & - & - & 0.5620 & 0.7380 & 0.3560\\
(12) & GPL~\cite{wang2021gpl} & - & - & - & - & - & - & 0.7400 & 0.3330 \\
(13) & miniLM reranker & $^\dagger$0.3901 & - & - & - & - & $^\ddag$0.5330 & $^\ddag$0.7570 & $^\ddag$0.3470 \\
(14) & monoT5-220M~\cite{nogueira2020document} & 0.3810 & 0.4909 & 0.7141 & 0.3279 & 0.5298 & 0.5674 & 0.7775 & 0.4136\\
(15) & monoT5-3B~\cite{nogueira2020document} & 0.3980 & \textbf{0.5281} & \textbf{0.7508} & 0.3876 & 0.6091 & \textbf{0.6334} & 0.7948 & \textbf{0.5137} \\
\midrule
& \multicolumn{5}{l}{\textit{InPars (ours)} [$\triangleright$ MARCO $\triangleright$ unsup in-domain]}\\
(16) & monoT5-3B & 0.3894 & 0.5087 & 0.7439 & \textbf{0.3967} & \textbf{0.6227} & 0.6297 & \textbf{0.8471} & 0.4987 \\

\bottomrule

\end{tabular}
}
\caption{Main results. Figures marked with a $\dagger$ and a $\ddag$ are from~\citet{reimers-2020-Curse_Dense_Retrieval} and \citet{thakur2021beir}, respectively. Experiments that are too expensive to run are marked with a \$ symbol. For example, evaluating Davinci as a reranker on the MS MARCO development set would cost approximately 6.3k USD. $\triangleright$ indicates the dataset used for finetuning the model.}
\label{tab:main}
\end{table*}

Robust04 contains long documents that would require large amounts of memory due to the quadratic cost of the Transformer models. Thus, during finetuning and inference, it is not possible to directly feed the entire text at once to our models. To address this issue, we use a slightly modified version of the MaxP technique of \citet{dai2019deeper}. We first segment each document into passages by applying a sliding window of 10 sentences with a stride of 5. We then obtain a relevance probability for each passage by classifying it independently. We select the highest probability among these passages as the relevance probability of the document; that is, we do not use the original (BM25) retrieval scores.

\noindent{\textbf{OpenAI Search API:}} We also experiment with OpenAI's Search API\footnote{https://beta.openai.com/docs/guides/search} as a reranker of 100 documents retrieved by BM25. The Search API provides an endpoint to perform semantic text search over a set of documents, where we provide a query and the top 100 documents retrieved by BM25, and the result is a reranked list with the respective scores. OpenAI does not disclose what is under the hood of their Search API, whether it was finetuned on IR datasets, the prompt used during inference, or its relation with cpt-text~\cite{neelakantan2022text}. Therefore we compare their Search API, which uses large LMs as rerankers, with our method, which uses the same LMs to train rerankers with lower inference cost.

We did not rerank more documents due to the high cost of the API (approximately 5.85 USD per 1000 query-document pairs of TREC-DL 2020 using Davinci). Nevertheless, we conducted a single experiment reranking 1000 documents per query using Curie and saw an improvement of almost two nDCG@10 points on TREC-DL 2020 in comparison to reranking 100 documents (0.5602 vs 0.5422). Thus, with more candidate documents, the figures reported for the other datasets in Table~\ref{tab:main} would probably be higher.

\section{Results}
\label{sec:results}

The main results are shown in Table~\ref{tab:main}. We observe that unsupervised models finetuned using our method (rows 7-8), outperform models of equivalent size available in OpenAI's Search API (rows 4-6).
For instance, our monoT5 with 3B parameters (row 8) is significantly better than the larger Curie and Davinci models (rows 5 and 6).
InPars also outperforms Contriever~\cite{izacard2021towards} and cpt-text~\cite{neelakantan2022text}, two unsupervised methods.
\citet{neelakantan2022text} (row 3) also use large LMs, but their numbers are much lower than those achieved by InPars or OpenAI's Search reranking documents from BM25 (rows 4-6).

We believe different factors contribute to these results:
1) Due to the query-document interaction, cross-encoders (rows 7-8) are more effective (and contextual) than the independent query and document encoding of bi-encoders (rows 3-6). This is inline with results from~\citet{thakur2021beir} which shows that bi-encoders and cross-encoders perform similarly when trained and evaluated on the same dataset, but cross-encoders show better zero-shot effectiveness. 2) The synthetic questions generated by InPars have closer resemblance to the queries that the retriever will see at inference time than the sentences extracted from texts used by Contriever and cpt-text.
3) The brand-new queries we generate combined with documents that the model otherwise will not use, provide a completely new training set with high diversity. Diversity in the training data has been shown to lead to improvement in performance of neural models~\cite{qu2021coda}.

\begin{table*}
\centering\centering\resizebox{1.0\textwidth}{!}{
\begin{tabular}{llllrrrrrrrH}
\toprule
& & & &\multicolumn{1}{c}{\textbf{MARCO}} & \multicolumn{2}{c}{\textbf{TREC-DL 2020}} & \multicolumn{2}{c}{\textbf{Robust04}} & \multicolumn{1}{c}{\textbf{NQ}} & \multicolumn{1}{c}{\textbf{TRECC}} & \multicolumn{1}{H}{\textbf{FiQA}}\\
& & \textbf{Input docs} & \textbf{Prompt} & \textbf{MRR@10} & \textbf{MAP}    & \small\textbf{nDCG@10} &  \textbf{MAP} & \small\textbf{nDCG@20} & \small\textbf{nDCG@10} & \small\textbf{nDCG@10} & \small\textbf{nDCG@10}\\
\midrule
(1) & monoT5-220M & Marco & Vanilla & 0.2585 & 0.3599 & 0.5764 & 0.2242 & 0.4017 & 0.3755 & 0.6727 & 0.3036  \\
(2) & monoT5-220M & In-domain & GBQ & 0.2279 & 0.3354 & 0.5451 & 0.2490 & 0.4268 & 0.3354 & 0.6666 & 0.3316\\
(3) & monoT5-3B & Marco & Vanilla & \textbf{0.2967} & \textbf{0.4334} & \textbf{0.6612} & \textbf{0.3397} & \textbf{0.5478} & 0.4870 & 0.7606 & ? \\
(4) & monoT5-3B & In-domain & GBQ & 0.2819 & 0.4100 & 0.6255 & 0.3180 & 0.5181 & \textbf{0.5134} & \textbf{0.7835} & \textbf{0.4103} \\
\bottomrule
\end{tabular}
}
\caption{Ablation: Few-shot results when using different input documents and prompts to generate questions.}
\label{tab:prompt_ablation}
\end{table*}



The last row (16) presents the results of monoT5-3B finetuned on MS MARCO and further finetuned on our synthetic datasets. Starting from the model from row 15, we finetuned them on questions generated from the in-domain collection.\footnote{TREC DL and MS MARCO results in rows 14-16 are from models finetuned for 10 epochs on MS MARCO. For the other datasets, the models were finetuned for one epoch on MS MARCO as this leads to better zero-shot effectiveness~\cite{Pradeep2020H2olooAT}.} This procedure achieves the state-of-the-art on Robust04 and TREC-COVID datasets, outperforming the zero-shot transfer achieved by models trained solely on MS MARCO (rows 9-15)\footnote{We do not report PARADE~\cite{li2020parade} results, that achieves an nDCG@20 of 0.6127 on Robust04, since it was finetuned on Robust04 data and are are only comparing against zero-shot models on this dataset.}. This result is especially noteworthy for TREC-COVID, in which the subject is relatively recent and unknown by many pretrained LMs. This indicates that the adaptation to the domain brought by our generation method led to large gains compared to the zero-shot approaches.

As for MS MARCO and TREC DL, further finetuning a model on 10k synthetic examples that was already trained on 530k manually annotated examples led to a slight decrease in effectiveness. We hypothesize that our synthetic examples are different from MS MARCO ones as we sample passages from the MS MARCO corpus to generate questions that do not often appear in the original relevant passage set (e.g., footnotes, texts about advertisements, passages with multiple topics, etc). Thus, any further training on different examples would be harmful when compared to training and evaluating on data of the same distribution.


\section{Ablation Study and Analysis}
In this section we take a deeper look into the properties of our proposed approach.

\subsection{Prompt Selection and Source Corpus}
\label{sec:prompt_ablation}
First, we investigate the impact of using different styles of prompts as well as using in-domain or ad hoc collections to generate synthetic data.
We compare IR metrics obtained with different prompt templates in Table~\ref{tab:prompt_ablation}. The column ``Input docs'' refers to the collection used to generate the questions.
Marco indicates that we sampled documents from the MS MARCO passage collection as input for generating questions since models trained on them have shown great generalization capabilities \cite{pradeep5h2oloo,thakur2021beir}.
In contrast, in-domain signifies that questions were generated from documents sampled from the same collections that the model is evaluated on. This approach can be regarded as zero-shot domain adaptation as we do not use any in-domain \textit{labeled} data to finetune our retrievers.
Since the MS MARCO development set and TREC-DL 2020 use the MS MARCO passage collection, this distinction of in-domain and Marco does not apply to these datasets.

Column ``Prompt'' in Table~\ref{tab:prompt_ablation} refers to the type of prompt fed to the language model during the question generation step. Preliminary experiments demonstrate that GBQ leads to better performance than the Vanilla prompt when used in combination with in-domain input documents. A marginal gain was observed when using the GBQ prompt on MS MARCO documents. Therefore, we combined Marco with the Vanilla prompt and in-domain with the GBQ prompt in our experiments.

It costs 100-300 USD to generate 100,000 questions using the Curie model regardless of the prompt type. Table~\ref{tab:examples_robust04} shows the difference in the predicted query for both prompts. It is notably observed that GBQ questions tend to be more descriptive and specific to the input document while Vanilla questions are more generic and can be answered by a broader set of documents.

Both MS MARCO and TREC-DL 2020 achieved the highest scores using the Vanilla prompt for both monoT5 220M and 3B models. However, for the other datasets (Robust04, NQ, and TREC-COVID), we obtained the highest scores with the GBQ prompt and using questions generated from their collections. The 3B models benefited the most from the in-domain examples, consistently outperforming the Marco results. For smaller monoT5 220M, there is no clear winner between the types of prompts and input documents.

\subsection{Model Size Impact on IR Metrics}
\label{sec:model_size}

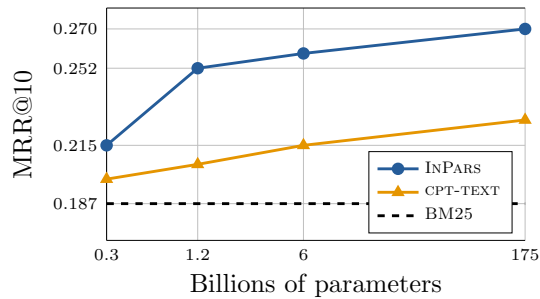
\begin{figure}
\centering
\begin{tikzpicture}[scale = 0.9]
\begin{axis}[
width=1.0\columnwidth,
height=0.65\columnwidth,
legend cell align=left,
mark options={mark size=3},
font=\scriptsize,
axis y line*=left,
xmin=0.3, xmax=175,
ymin=0.17, ymax=0.28,
xmode=log,
log ticks with fixed point,
xtick={0.3, 1.2, 6, 175},
ytick={0.1874, 0.215, 0.2515, 0.2701},
legend pos=south east,
y tick label style={
    /pgf/number format/.cd,
        fixed,
        fixed zerofill,
        precision=3,
    /tikz/.cd
},
ytick style={draw=none},
xtick style={draw=none},
xmajorgrids=true,
ymajorgrids=true,
xlabel style={font = \normalsize, yshift=1ex},
xlabel=Billions of parameters,
ylabel= MRR@10,
ylabel style={font = \normalsize, yshift=0ex}]
    \addplot[mark=*, very thick, blu, mark options={scale=1}] plot coordinates {
    (0.3, 0.2150)(1.2, 0.2515)(6, 0.2585)(175, 0.2701)
    };
    \addlegendentry{\textsc{InPars}}
    
\addplot[mark=triangle*, very thick, orng, mark options={scale=1}] plot coordinates {
    (0.3, 0.199)(1.2, 0.206)(6, 0.215)(175, 0.227)
    };
    \addlegendentry{\textsc{cpt-text}}
    
    \addplot[very thick, dashed, black, mark options={scale=1}] plot coordinates {
    (0.3, 0.1874)(175, 0.1874)
    };
    \addlegendentry{\textsc{BM25}}
\end{axis}
\end{tikzpicture}
\caption{MRR@10 on the MS MARCO development set achieved by InPars using monoT5-220M reranker trained on synthetic questions generated by GPT-3 models of different sizes. Figures for cpt-text are from~\cite{neelakantan2022text}. Note the log scale for the x-axis.}
\label{fig:mrr_vs_usd}
\end{figure}

In Figure~\ref{fig:mrr_vs_usd} we present the effectiveness on the MS MARCO development set of the monoT5-220M reranker trained on synthetic questions generated using different sizes of GPT-3. As we increase the model size, the IR metric keeps increasing, although very slowly. A hypothesis that explains this gain is that larger models can generate more relevant questions to a given passage, instead of broad and more general questions. Using more specific questions avoids confusing the reranker with questions that are relevant to many passages, which can lead to false-negative examples.


This relation between GPT-3 model sizes and their IR effectiveness is also observed in the OpenAI Search results presented in Table~\ref{tab:main}. However, in those results, the Davinci model rivals Curie as the best performing, with the latter delivering a higher score on most experiments.

\subsection{Filtering by the Most Likely Questions}
\label{sec:filtering_ablation}
As mentioned in Section~\ref{sec:method}, we used only the top $K=10,000$ pairs w.r.t to $p_q$ to finetune the rerankers. 
Finetuning on all 100,000 synthetic examples leads to a 4 MRR@10 points decrease on MS MARCO when compared to the top $K$ filtering approach. Thus, to avoid relying on the test sets and keep the method fully unsupervised, we only evaluated this difference on MS MARCO and applied the filtering approach to all other datasets.

\subsection{Was GPT-3 Trained on Supervised IR Data?}

The answer to this question is not disclosed to the public, but we tried to answer it by measuring the number of questions produced by GPT-3 that match those in the MS MARCO dataset. Of the 93,200 unique questions produced by the largest model (Davinci), 5,285 were also found in the 1M questions from MS MARCO, i.e., less than 5.7\%. Most of these questions contain a single entity (e.g., ``what is freedom of the press'', ``what is the fda'', ``who is gary young'') that are likely to be answered by multiple documents in the collection.
Also, the Davinci model produced questions from 6,396 documents marked as relevant in either the training or development sets of MS MARCO. Of those synthetic questions, 725 were found in MS MARCO ground-truth, i.e., less than 12\%.
Similar numbers were found for smaller models.
We argue that these low percentages are evidence that the models were not finetuned on MS MARCO, or at least, they did not memorize it.

\section{Conclusion and Future Work}
 
In this work, we presented \textit{InPars}, a method to generate synthetic training data for IR tasks using large LMs in a few-shot manner. 
This allows one to harness the information learned by large models in a more efficient and effective way.

Our experiments demonstrated that using large LMs to generate synthetic training data is a promising direction for development of neural retrievers.
There are, however, many directions not explored in this work that we leave as future work: 1) Finetuning dense retrievers on our synthetic data; 2) Using ``bad questions'' as negative training examples; 3) Scale up our synthetic datasets to millions of examples; 4) More sophisticated methods to select (question, relevant document) pairs.

\section*{Acknowledgments}
This research was partially funded by a grant from Fundação de Amparo à Pesquisa do Estado de São Paulo (FAPESP)
2020/09753-5.
We also would like to thank Google Cloud for credits to support this work.

\bibliography{main}

\begin{thebibliography}{54}
\expandafter\ifx\csname natexlab\endcsname\relax\def\natexlab#1{#1}\fi

\bibitem[{Anaby-Tavor et~al.(2020)Anaby-Tavor, Carmeli, Goldbraich, Kantor,
  Kour, Shlomov, Tepper, and Zwerdling}]{anaby2020not}
Ateret Anaby-Tavor, Boaz Carmeli, Esther Goldbraich, Amir Kantor, George Kour,
  Segev Shlomov, Naama Tepper, and Naama Zwerdling. 2020.
\newblock Do not have enough data? deep learning to the rescue!
\newblock In \emph{Proceedings of the AAAI Conference on Artificial
  Intelligence}, volume~34, pages 7383--7390.

\bibitem[{Brown et~al.(2020)Brown, Mann, Ryder, Subbiah, Kaplan, Dhariwal,
  Neelakantan, Shyam, Sastry, Askell, Agarwal, Herbert{-}Voss, Krueger,
  Henighan, Child, Ramesh, Ziegler, Wu, Winter, Hesse, Chen, Sigler, Litwin,
  Gray, Chess, Clark, Berner, McCandlish, Radford, Sutskever, and
  Amodei}]{DBLP:journals/corr/abs-2005-14165}
Tom~B. Brown, Benjamin Mann, Nick Ryder, Melanie Subbiah, Jared Kaplan,
  Prafulla Dhariwal, Arvind Neelakantan, Pranav Shyam, Girish Sastry, Amanda
  Askell, Sandhini Agarwal, Ariel Herbert{-}Voss, Gretchen Krueger, Tom
  Henighan, Rewon Child, Aditya Ramesh, Daniel~M. Ziegler, Jeffrey Wu, Clemens
  Winter, Christopher Hesse, Mark Chen, Eric Sigler, Mateusz Litwin, Scott
  Gray, Benjamin Chess, Jack Clark, Christopher Berner, Sam McCandlish, Alec
  Radford, Ilya Sutskever, and Dario Amodei. 2020.
\newblock \href {http://arxiv.org/abs/2005.14165} {Language models are few-shot
  learners}.
\newblock \emph{CoRR}, abs/2005.14165.

\bibitem[{Chen et~al.(2017)Chen, Gallagher, Blanco, and
  Culpepper}]{chen2017efficient}
Ruey-Cheng Chen, Luke Gallagher, Roi Blanco, and J~Shane Culpepper. 2017.
\newblock Efficient cost-aware cascade ranking in multi-stage retrieval.
\newblock In \emph{Proceedings of the 40th International ACM SIGIR Conference
  on Research and Development in Information Retrieval}, pages 445--454.

\bibitem[{Craswell et~al.(2021)Craswell, Mitra, Yilmaz, and
  Campos}]{DBLP:journals/corr/abs-2102-07662}
Nick Craswell, Bhaskar Mitra, Emine Yilmaz, and Daniel Campos. 2021.
\newblock \href {http://arxiv.org/abs/2102.07662} {Overview of the {TREC} 2020
  deep learning track}.
\newblock \emph{CoRR}, abs/2102.07662.

\bibitem[{Dai and Callan(2019)}]{dai2019deeper}
Zhuyun Dai and Jamie Callan. 2019.
\newblock Deeper text understanding for ir with contextual neural language
  modeling.
\newblock In \emph{Proceedings of the 42nd International ACM SIGIR Conference
  on Research and Development in Information Retrieval}, pages 985--988.

\bibitem[{Fadaee et~al.(2017)Fadaee, Bisazza, and Monz}]{fadaee-etal-2017-data}
Marzieh Fadaee, Arianna Bisazza, and Christof Monz. 2017.
\newblock \href {https://doi.org/10.18653/v1/P17-2090} {Data augmentation for
  low-resource neural machine translation}.
\newblock In \emph{Proceedings of the 55th Annual Meeting of the Association
  for Computational Linguistics (Volume 2: Short Papers)}, pages 567--573,
  Vancouver, Canada. Association for Computational Linguistics.

\bibitem[{Gao and Callan(2021)}]{gao2021unsupervised}
Luyu Gao and Jamie Callan. 2021.
\newblock Unsupervised corpus aware language model pre-training for dense
  passage retrieval.
\newblock \emph{arXiv preprint arXiv:2108.05540}.

\bibitem[{Han et~al.(2021)Han, Babuschkin, Edwards, Neelakantan, Xu, Polu, Ray,
  Shyam, Ramesh, Radford et~al.}]{han2021unsupervised}
Jesse~Michael Han, Igor Babuschkin, Harrison Edwards, Arvind Neelakantan, Tao
  Xu, Stanislas Polu, Alex Ray, Pranav Shyam, Aditya Ramesh, Alec Radford,
  et~al. 2021.
\newblock Unsupervised neural machine translation with generative language
  models only.
\newblock \emph{arXiv preprint arXiv:2110.05448}.

\bibitem[{Holtzman et~al.(2019)Holtzman, Buys, Du, Forbes, and
  Choi}]{holtzman2019curious}
Ari Holtzman, Jan Buys, Li~Du, Maxwell Forbes, and Yejin Choi. 2019.
\newblock The curious case of neural text degeneration.
\newblock In \emph{International Conference on Learning Representations}.

\bibitem[{Izacard et~al.(2021)Izacard, Caron, Hosseini, Riedel, Bojanowski,
  Joulin, and Grave}]{izacard2021towards}
Gautier Izacard, Mathilde Caron, Lucas Hosseini, Sebastian Riedel, Piotr
  Bojanowski, Armand Joulin, and Edouard Grave. 2021.
\newblock Towards unsupervised dense information retrieval with contrastive
  learning.
\newblock \emph{arXiv preprint arXiv:2112.09118}.

\bibitem[{Johnson et~al.(2019)Johnson, Douze, and
  J{\'e}gou}]{johnson2019billion}
Jeff Johnson, Matthijs Douze, and Herv{\'e} J{\'e}gou. 2019.
\newblock Billion-scale similarity search with gpus.
\newblock \emph{IEEE Transactions on Big Data}.

\bibitem[{Karpukhin et~al.(2020)Karpukhin, Oguz, Min, Wu, Edunov, Chen, and
  Yih}]{DBLP:journals/corr/abs-2004-04906}
Vladimir Karpukhin, Barlas Oguz, Sewon Min, Ledell Wu, Sergey Edunov, Danqi
  Chen, and Wen{-}tau Yih. 2020.
\newblock \href {http://arxiv.org/abs/2004.04906} {Dense passage retrieval for
  open-domain question answering}.
\newblock \emph{CoRR}, abs/2004.04906.

\bibitem[{Khattab and Zaharia(2020)}]{DBLP:journals/corr/abs-2004-12832}
Omar Khattab and Matei Zaharia. 2020.
\newblock \href {http://arxiv.org/abs/2004.12832} {Colbert: Efficient and
  effective passage search via contextualized late interaction over {BERT}}.
\newblock \emph{CoRR}, abs/2004.12832.

\bibitem[{Kobayashi(2018)}]{kobayashi2018contextual}
Sosuke Kobayashi. 2018.
\newblock Contextual augmentation: Data augmentation by words with paradigmatic
  relations.
\newblock \emph{arXiv preprint arXiv:1805.06201}.

\bibitem[{Kumar et~al.(2020)Kumar, Choudhary, and Cho}]{kumar2020data}
Varun Kumar, Ashutosh Choudhary, and Eunah Cho. 2020.
\newblock Data augmentation using pre-trained transformer models.
\newblock In \emph{Proceedings of the 2nd Workshop on Life-long Learning for
  Spoken Language Systems}, pages 18--26.

\bibitem[{Kwiatkowski et~al.(2019)Kwiatkowski, Palomaki, Redfield, Collins,
  Parikh, Alberti, Epstein, Polosukhin, Kelcey, Devlin, Lee, Toutanova, Jones,
  Chang, Dai, Uszkoreit, Le, and Petrov}]{47761}
Tom Kwiatkowski, Jennimaria Palomaki, Olivia Redfield, Michael Collins, Ankur
  Parikh, Chris Alberti, Danielle Epstein, Illia Polosukhin, Matthew Kelcey,
  Jacob Devlin, Kenton Lee, Kristina~N. Toutanova, Llion Jones, Ming-Wei Chang,
  Andrew Dai, Jakob Uszkoreit, Quoc Le, and Slav Petrov. 2019.
\newblock Natural questions: a benchmark for question answering research.
\newblock \emph{Transactions of the Association of Computational Linguistics}.

\bibitem[{Li et~al.(2020)Li, Yates, MacAvaney, He, and Sun}]{li2020parade}
Canjia Li, Andrew Yates, Sean MacAvaney, Ben He, and Yingfei Sun. 2020.
\newblock Parade: Passage representation aggregation for document reranking.
\newblock \emph{arXiv preprint arXiv:2008.09093}.

\bibitem[{Lin et~al.(2021)Lin, Ma, Lin, Yang, Pradeep, and
  Nogueira}]{lin2021pyserini}
Jimmy Lin, Xueguang Ma, Sheng-Chieh Lin, Jheng-Hong Yang, Ronak Pradeep, and
  Rodrigo Nogueira. 2021.
\newblock Pyserini: A python toolkit for reproducible information retrieval
  research with sparse and dense representations.
\newblock In \emph{Proceedings of the 44th Annual International ACM SIGIR
  Conference on Research and Development in Information Retrieval (SIGIR
  2021)}.

\bibitem[{Liu et~al.(2022)Liu, Swayamdipta, Smith, and Choi}]{liu2022wanli}
Alisa Liu, Swabha Swayamdipta, Noah~A. Smith, and Yejin Choi. 2022.
\newblock \href {http://arxiv.org/abs/2201.05955} {Wanli: Worker and ai
  collaboration for natural language inference dataset creation}.

\bibitem[{Liu et~al.(2017)Liu, Xiao, Ou, and Si}]{liu2017cascade}
Shichen Liu, Fei Xiao, Wenwu Ou, and Luo Si. 2017.
\newblock Cascade ranking for operational e-commerce search.
\newblock In \emph{Proceedings of the 23rd ACM SIGKDD International Conference
  on Knowledge Discovery and Data Mining}, pages 1557--1565.

\bibitem[{Ma et~al.(2021)Ma, Korotkov, Yang, Hall, and
  McDonald}]{ma-etal-2021-zero}
Ji~Ma, Ivan Korotkov, Yinfei Yang, Keith Hall, and Ryan McDonald. 2021.
\newblock \href {https://doi.org/10.18653/v1/2021.eacl-main.92} {Zero-shot
  neural passage retrieval via domain-targeted synthetic question generation}.
\newblock In \emph{Proceedings of the 16th Conference of the European Chapter
  of the Association for Computational Linguistics: Main Volume}, pages
  1075--1088, Online. Association for Computational Linguistics.

\bibitem[{Matveeva et~al.(2006)Matveeva, Burges, Burkard, Laucius, and
  Wong}]{matveeva2006high}
Irina Matveeva, Chris Burges, Timo Burkard, Andy Laucius, and Leon Wong. 2006.
\newblock High accuracy retrieval with multiple nested ranker.
\newblock In \emph{Proceedings of the 29th annual international ACM SIGIR
  conference on Research and development in information retrieval}, pages
  437--444.

\bibitem[{Meng et~al.(2022)Meng, Huang, Zhang, and Han}]{meng2022generating}
Yu~Meng, Jiaxin Huang, Yu~Zhang, and Jiawei Han. 2022.
\newblock \href {http://arxiv.org/abs/2202.04538} {Generating training data
  with language models: Towards zero-shot language understanding}.

\bibitem[{Mohapatra et~al.(2021)Mohapatra, Pandey, Contractor, and
  Joshi}]{mohapatra2021simulated}
Biswesh Mohapatra, Gaurav Pandey, Danish Contractor, and Sachindra Joshi. 2021.
\newblock Simulated chats for building dialog systems: Learning to generate
  conversations from instructions.
\newblock In \emph{Findings of the Association for Computational Linguistics:
  EMNLP 2021}, pages 1190--1203.

\bibitem[{Neelakantan et~al.(2022)Neelakantan, Xu, Puri, Radford, Han, Tworek,
  Yuan, Tezak, Kim, Hallacy et~al.}]{neelakantan2022text}
Arvind Neelakantan, Tao Xu, Raul Puri, Alec Radford, Jesse~Michael Han, Jerry
  Tworek, Qiming Yuan, Nikolas Tezak, Jong~Wook Kim, Chris Hallacy, et~al.
  2022.
\newblock Text and code embeddings by contrastive pre-training.
\newblock \emph{arXiv preprint arXiv:2201.10005}.

\bibitem[{Nguyen et~al.(2016)Nguyen, Rosenberg, Song, Gao, Tiwary, Majumder,
  and Deng}]{DBLP:journals/corr/NguyenRSGTMD16}
Tri Nguyen, Mir Rosenberg, Xia Song, Jianfeng Gao, Saurabh Tiwary, Rangan
  Majumder, and Li~Deng. 2016.
\newblock \href {http://arxiv.org/abs/1611.09268} {{MS} {MARCO:} {A} human
  generated machine reading comprehension dataset}.
\newblock \emph{CoRR}, abs/1611.09268.

\bibitem[{Nogueira et~al.(2020)Nogueira, Jiang, Pradeep, and
  Lin}]{nogueira2020document}
Rodrigo Nogueira, Zhiying Jiang, Ronak Pradeep, and Jimmy Lin. 2020.
\newblock Document ranking with a pretrained sequence-to-sequence model.
\newblock In \emph{Proceedings of the 2020 Conference on Empirical Methods in
  Natural Language Processing: Findings}, pages 708--718.

\bibitem[{Papanikolaou and Pierleoni(2020)}]{papanikolaou2020dare}
Yannis Papanikolaou and Andrea Pierleoni. 2020.
\newblock Dare: Data augmented relation extraction with gpt-2.
\newblock \emph{arXiv preprint arXiv:2004.13845}.

\bibitem[{Pradeep et~al.(2020{\natexlab{a}})Pradeep, Ma, Zhang, Cui, Xu,
  Nogueira, Lin, and Cheriton}]{Pradeep2020H2olooAT}
Ronak Pradeep, Xueguang Ma, Xinyu Zhang, H.~Cui, Ruizhou Xu, Rodrigo Nogueira,
  Jimmy~J. Lin, and D.~Cheriton. 2020{\natexlab{a}}.
\newblock H2oloo at trec 2020: When all you got is a hammer... deep learning,
  health misinformation, and precision medicine.
\newblock In \emph{TREC}.

\bibitem[{Pradeep et~al.(2020{\natexlab{b}})Pradeep, Ma, Zhang, Cui, Xu,
  Nogueira, and Lin}]{pradeep5h2oloo}
Ronak Pradeep, Xueguang Ma, Xinyu Zhang, Hang Cui, Ruizhou Xu, Rodrigo
  Nogueira, and Jimmy Lin. 2020{\natexlab{b}}.
\newblock H2oloo at trec 2020: When all you got is a hammer... deep learning,
  health misinformation, and precision medicine.
\newblock \emph{Corpus}, 5(d3):d2.

\bibitem[{Pradeep et~al.(2021)Pradeep, Nogueira, and Lin}]{pradeep2021expando}
Ronak Pradeep, Rodrigo Nogueira, and Jimmy Lin. 2021.
\newblock The expando-mono-duo design pattern for text ranking with pretrained
  sequence-to-sequence models.
\newblock \emph{arXiv preprint arXiv:2101.05667}.

\bibitem[{Qu et~al.(2021)Qu, Shen, Shen, Sajeev, Chen, and Han}]{qu2021coda}
Yanru Qu, Dinghan Shen, Yelong Shen, Sandra Sajeev, Weizhu Chen, and Jiawei
  Han. 2021.
\newblock \href {https://openreview.net/forum?id=Ozk9MrX1hvA} {Co{\{}da{\}}:
  Contrast-enhanced and diversity-promoting data augmentation for natural
  language understanding}.
\newblock In \emph{International Conference on Learning Representations}.

\bibitem[{Rae et~al.(2021)Rae, Borgeaud, Cai, Millican, Hoffmann, Song,
  Aslanides, Henderson, Ring, Young et~al.}]{rae2021scaling}
Jack~W Rae, Sebastian Borgeaud, Trevor Cai, Katie Millican, Jordan Hoffmann,
  Francis Song, John Aslanides, Sarah Henderson, Roman Ring, Susannah Young,
  et~al. 2021.
\newblock Scaling language models: Methods, analysis \& insights from training
  gopher.
\newblock \emph{arXiv preprint arXiv:2112.11446}.

\bibitem[{Raffel et~al.(2020)Raffel, Shazeer, Roberts, Lee, Narang, Matena,
  Zhou, Li, and Liu}]{raffel2020exploring}
Colin Raffel, Noam Shazeer, Adam Roberts, Katherine Lee, Sharan Narang, Michael
  Matena, Yanqi Zhou, Wei Li, and Peter~J Liu. 2020.
\newblock Exploring the limits of transfer learning with a unified text-to-text
  transformer.
\newblock \emph{Journal of Machine Learning Research}, 21:1--67.

\bibitem[{Ram et~al.(2021)Ram, Shachaf, Levy, Berant, and
  Globerson}]{ram2021learning}
Ori Ram, Gal Shachaf, Omer Levy, Jonathan Berant, and Amir Globerson. 2021.
\newblock Learning to retrieve passages without supervision.
\newblock \emph{arXiv preprint arXiv:2112.07708}.

\bibitem[{Reimers and Gurevych(2020)}]{reimers-2020-Curse_Dense_Retrieval}
Nils Reimers and Iryna Gurevych. 2020.
\newblock \href {https://arxiv.org/abs/2012.14210} {The curse of dense
  low-dimensional information retrieval for large index sizes}.
\newblock \emph{arXiv preprint arXiv:2012.14210}.

\bibitem[{Roberts et~al.(2020)Roberts, Alam, Bedrick, Demner-Fushman, Lo,
  Soboroff, Voorhees, Wang, and Hersh}]{10.1093/jamia/ocaa091}
Kirk Roberts, Tasmeer Alam, Steven Bedrick, Dina Demner-Fushman, Kyle Lo, Ian
  Soboroff, Ellen Voorhees, Lucy~Lu Wang, and William~R Hersh. 2020.
\newblock \href {https://doi.org/10.1093/jamia/ocaa091} {{TREC-COVID: rationale
  and structure of an information retrieval shared task for COVID-19}}.
\newblock \emph{Journal of the American Medical Informatics Association},
  27(9):1431--1436.

\bibitem[{ROBERTSON et~al.(1995)ROBERTSON, WALKER, JONES, HANCOCK-BEAULIEU, and
  GATFORD}]{robertson1995okapi}
SE~ROBERTSON, S~WALKER, S~JONES, MM~HANCOCK-BEAULIEU, and M~GATFORD. 1995.
\newblock Okapi at trec-3.
\newblock \emph{NIST special publication}, (500225):109--123.

\bibitem[{Sanh et~al.(2021)Sanh, Webson, Raffel, Bach, Sutawika, Alyafeai,
  Chaffin, Stiegler, Scao, Raja, Dey, Bari, Xu, Thakker, Sharma, Szczechla,
  Kim, Chhablani, Nayak, Datta, Chang, Jiang, Wang, Manica, Shen, Yong, Pandey,
  Bawden, Wang, Neeraj, Rozen, Sharma, Santilli, Fevry, Fries, Teehan,
  Biderman, Gao, Bers, Wolf, and Rush}]{sanh2021multitask}
Victor Sanh, Albert Webson, Colin Raffel, Stephen~H. Bach, Lintang Sutawika,
  Zaid Alyafeai, Antoine Chaffin, Arnaud Stiegler, Teven~Le Scao, Arun Raja,
  Manan Dey, M~Saiful Bari, Canwen Xu, Urmish Thakker, Shanya~Sharma Sharma,
  Eliza Szczechla, Taewoon Kim, Gunjan Chhablani, Nihal Nayak, Debajyoti Datta,
  Jonathan Chang, Mike Tian-Jian Jiang, Han Wang, Matteo Manica, Sheng Shen,
  Zheng~Xin Yong, Harshit Pandey, Rachel Bawden, Thomas Wang, Trishala Neeraj,
  Jos Rozen, Abheesht Sharma, Andrea Santilli, Thibault Fevry, Jason~Alan
  Fries, Ryan Teehan, Stella Biderman, Leo Gao, Tali Bers, Thomas Wolf, and
  Alexander~M. Rush. 2021.
\newblock \href {http://arxiv.org/abs/2110.08207} {Multitask prompted training
  enables zero-shot task generalization}.

\bibitem[{Santhanam et~al.(2021)Santhanam, Khattab, Saad-Falcon, Potts, and
  Zaharia}]{santhanam2021colbertv2}
Keshav Santhanam, Omar Khattab, Jon Saad-Falcon, Christopher Potts, and Matei
  Zaharia. 2021.
\newblock Colbertv2: Effective and efficient retrieval via lightweight late
  interaction.
\newblock \emph{arXiv preprint arXiv:2112.01488}.

\bibitem[{Schick and Sch{\"u}tze(2021{\natexlab{a}})}]{schick2021generating}
Timo Schick and Hinrich Sch{\"u}tze. 2021{\natexlab{a}}.
\newblock Generating datasets with pretrained language models.
\newblock \emph{arXiv preprint arXiv:2104.07540}.

\bibitem[{Schick and
  Sch{\"u}tze(2021{\natexlab{b}})}]{schick-schutze-2021-just}
Timo Schick and Hinrich Sch{\"u}tze. 2021{\natexlab{b}}.
\newblock \href {https://doi.org/10.18653/v1/2021.naacl-main.185} {It{'}s not
  just size that matters: Small language models are also few-shot learners}.
\newblock In \emph{Proceedings of the 2021 Conference of the North American
  Chapter of the Association for Computational Linguistics: Human Language
  Technologies}, pages 2339--2352, Online. Association for Computational
  Linguistics.

\bibitem[{Sharami et~al.(2022)Sharami, Shterionov, and
  Spronck}]{sharami2022selecting}
Javad Pourmostafa~Roshan Sharami, Dimitar Shterionov, and Pieter Spronck. 2022.
\newblock \href {http://arxiv.org/abs/2112.06096} {Selecting parallel in-domain
  sentences for neural machine translation using monolingual texts}.

\bibitem[{Thakur et~al.(2021)Thakur, Reimers, R{\"u}ckl{\'e}, Srivastava, and
  Gurevych}]{thakur2021beir}
Nandan Thakur, Nils Reimers, Andreas R{\"u}ckl{\'e}, Abhishek Srivastava, and
  Iryna Gurevych. 2021.
\newblock \href {https://openreview.net/forum?id=wCu6T5xFjeJ} {{BEIR}: A
  heterogeneous benchmark for zero-shot evaluation of information retrieval
  models}.
\newblock In \emph{Thirty-fifth Conference on Neural Information Processing
  Systems Datasets and Benchmarks Track (Round 2)}.

\bibitem[{Voorhees(2005)}]{96071}
Ellen Voorhees. 2005.
\newblock \href {https://doi.org/https://doi.org/10.6028/NIST.SP.500-261}
  {Overview of the trec 2004 robust retrieval track}.

\bibitem[{Voorhees(2004)}]{Voorhees2004OverviewOT}
Ellen~M. Voorhees. 2004.
\newblock Overview of trec 2004.
\newblock In \emph{TREC}.

\bibitem[{Voorhees and Harman(1999)}]{Voorhees1999OverviewOT}
Ellen~M. Voorhees and Donna~K. Harman. 1999.
\newblock Overview of the eighth text retrieval conference (trec-8).
\newblock In \emph{TREC}.

\bibitem[{Wang et~al.(2021)Wang, Thakur, Reimers, and Gurevych}]{wang2021gpl}
Kexin Wang, Nandan Thakur, Nils Reimers, and Iryna Gurevych. 2021.
\newblock Gpl: Generative pseudo labeling for unsupervised domain adaptation of
  dense retrieval.
\newblock \emph{arXiv preprint arXiv:2112.07577}.

\bibitem[{Wang et~al.(2011)Wang, Lin, and Metzler}]{wang2011cascade}
Lidan Wang, Jimmy Lin, and Donald Metzler. 2011.
\newblock A cascade ranking model for efficient ranked retrieval.
\newblock In \emph{Proceedings of the 34th international ACM SIGIR conference
  on Research and development in Information Retrieval}, pages 105--114.

\bibitem[{Wei et~al.(2022)Wei, Bosma, Zhao, Guu, Yu, Lester, Du, Dai, and
  Le}]{wei2022finetuned}
Jason Wei, Maarten Bosma, Vincent Zhao, Kelvin Guu, Adams~Wei Yu, Brian Lester,
  Nan Du, Andrew~M. Dai, and Quoc~V Le. 2022.
\newblock \href {https://openreview.net/forum?id=gEZrGCozdqR} {Finetuned
  language models are zero-shot learners}.
\newblock In \emph{International Conference on Learning Representations}.

\bibitem[{Winata et~al.(2021)Winata, Madotto, Lin, Liu, Yosinski, and
  Fung}]{winata-etal-2021-language}
Genta~Indra Winata, Andrea Madotto, Zhaojiang Lin, Rosanne Liu, Jason Yosinski,
  and Pascale Fung. 2021.
\newblock \href {https://doi.org/10.18653/v1/2021.mrl-1.1} {Language models are
  few-shot multilingual learners}.
\newblock In \emph{Proceedings of the 1st Workshop on Multilingual
  Representation Learning}, pages 1--15, Punta Cana, Dominican Republic.
  Association for Computational Linguistics.

\bibitem[{Yang et~al.(2020)Yang, Malaviya, Fernandez, Swayamdipta, Le~Bras,
  Wang, Bhagavatula, Choi, and Downey}]{yang2020g}
Yiben Yang, Chaitanya Malaviya, Jared Fernandez, Swabha Swayamdipta, Ronan
  Le~Bras, Ji-Ping Wang, Chandra Bhagavatula, Yejin Choi, and Doug Downey.
  2020.
\newblock G-daug: Generative data augmentation for commonsense reasoning.
\newblock In \emph{Proceedings of the 2020 Conference on Empirical Methods in
  Natural Language Processing: Findings}, pages 1008--1025.

\bibitem[{Yilmaz et~al.(2020)Yilmaz, Craswell, Mitra, and
  Campos}]{yilmaz2020reliability}
Emine Yilmaz, Nick Craswell, Bhaskar Mitra, and Daniel Campos. 2020.
\newblock On the reliability of test collections for evaluating systems of
  different types.
\newblock In \emph{proceedings of the 43rd International ACM SIGIR Conference
  on Research and Development in Information Retrieval}, pages 2101--2104.

\bibitem[{Yu et~al.(2021)Yu, Liu, and Fung}]{yu2021adaptsum}
Tiezheng Yu, Zihan Liu, and Pascale Fung. 2021.
\newblock \href {http://arxiv.org/abs/2103.11332} {Adaptsum: Towards
  low-resource domain adaptation for abstractive summarization}.

\end{thebibliography}
\bibliographystyle{acl_natbib}

\newpage
\appendix
\section{Appendix}

\begin{table*}[htbp!]
 \begin{small}
\setlength\tabcolsep{6pt} 
 \caption{Examples of questions generated from the Robust04 dataset using the Curie model with Vanilla and GBQ prompts.}
 \label{tab:examples_robust04}
\begin{tabularx}{\textwidth}{X}
\toprule
\textbf{\textcolor{blu}{Document:}} {\footnotesize South Korean President Kim Yong-sam said Saturday it is imperative that Japan and South Korea work to end North Korea's isolation and halt its suspected nuclear arms program. In a joint press conference with Prime Minister Morihiro Hosokawa after a second round of summit talks, Kim said that South Korea and Japan share a common interest in developing economic cooperation in the Asia-Pacific region. Peace and security on the Korean peninsula are essential to stability in the region, but vestiges of the Cold War still linger on the peninsula, Kim said. Kim stated that it is imperative that neighboring countries work together toward making North Korea discard its isolationist policy. He also said it is imperative that North Korea give up its suspected nuclear arms development program and return to the South-North dialogue. Reporting his agreement with Hosokawa on economic cooperation in the East Asian region, Kim noted that both sides agreed the lopsided bilateral trade imbalance must be resolved. The president ends a three-day visit to Japan Saturday afternoon, leaving for Shanghai for talks with Chinese leaders. Kim said it was the insincerity of North Korea that brought about the referral of the nuclear issue to the United Nations Security Council. Measures should be taken in accordance with the North's attitude, the president said, suggesting a flexible stance toward Pyongyang. Hosokawa stressed the need for a firm and unified response of international society to the issue while leaving open a window for dialogue with North Korea.}   \\
\textbf{\textcolor{orng}{Vanilla question:}} \textit{What is the president's position on North Korea?}  \\ 
\textbf{\textcolor{rd}{GBQ question:}} \textit{What is the South Korean president's position on North Korea?}  \\
  \midrule
\textbf{\textcolor{blu}{Document:}} {\footnotesize John West has resigned as non-executive chairman of Dalgety, the food group, and of Bridon, maker of wire rope and engineered products, having just suffered a stroke. Dalgety has appointed Maurice Warren, chief executive, to succeed West, 65, who became chairman of Dalgety only last September. Richard Clothier, who had been due to become chief executive when Warren turns 60 in June, will now do so on April 1. Warren has agreed to remain chairman for the indefinite future and says the appointment will not affect his plans to become non-executive chairman of the South West Electricity Board in June. Bridon yesterday named Derek Edwards, a non-executive director of the company for the past eight years, as chairman, and Brian Clayton as chief executive. Clayton has been responsible for day-to-day executive decisions since David Allday resigned as chief executive in September. John Hogan has been appointed chief operating officer of Lasmo, the independent oil exploration and production company. He replaces Joe Darby who recently became chief executive after Chris Greentree decided to step down. John Hogan has been managing director of Lasmo's North Sea operations for the past four years. At 39, he is generally regarded as one of the industry's younger generation of pragmatic managers who have to weigh more keenly the financial risks and rewards of oil exploration in a climate of persistently low oil prices.} \\
Vanilla question: \textit{What is the name of the food group and the wire rope maker?}  \\ 
GBQ question:  \textit{Information on John West's resignation as non-executive chairman of Dalgety and Bridon}  \\
\midrule
\textbf{\textcolor{blu}{Document:}} {\footnotesize WIDE PRICE fluctuations in many commodity markets this week resulted from relatively modest buying and selling orders in thin, pre-Christmas conditions. The sharpest movement was gold's Dollars 4.85-a-troy-ounce drop over Monday and Tuesday, which traders said was initiated by selling from an individual US trade house. As the price slid towards a support level at Dollars 332 an ounce one dealer told the Reuter news agency that 'under normal circumstances the volume traded would not have moved the market so much'. The gold price was steadier yesterday, gaining 55 cents to Dollars 332.85 an ounce, but was still Dollars 4.30 down on the week so far. The platinum price followed a similar pattern, recovering Dollars 1.10 at yesterday's afternoon fixing to reach Dollars 359.60 an ounce, Dollars 3.90 below last Friday's level, while cash silver was six cents down overall at 370.5 cents an ounce. Most base metals markets at the London Metal Exchange presented a mirror image of the movements in the gold market, slipping back a little yesterday following strong gains earlier in the week. Copper was a case in point. Modest Chinese buying and worries about copper workers joining a Polish general strike were less influential than technical factors in the three months delivery position's Pounds 54.50 rise to Pounds 1,489.25 a tonne over the first two days, dealers said. Mr Ted Arnold, analyst at the Merrill Lynch financial services group, pointed to the profound effect commodity funds, managing about Dollars 26bn, were having on traded metals markets.' These funds tend to work primarily on technical analysis,' he said.} \\
\textbf{\textcolor{orng}{Vanilla question:}} \textit{What is the price of gold?}  \\ 
\textbf{\textcolor{rd}{GBQ question:}}  \textit{What is the effect of commodity funds on the price of metals?}  \\
\bottomrule
\end{tabularx}

 \end{small}
\end{table*}

\begin{table*}[htbp!]
 \begin{small}
\setlength\tabcolsep{6pt}
 \caption{Examples of questions generated from the TREC-COVID dataset using the Curie model with generated good and bad questions.}
 \label{tab:examples_trec_covid}
\begin{tabularx}{\textwidth}{X}
\toprule
\textbf{\textcolor{blu}{Document:}} {\footnotesize [Ethical principles compromised during the COVID-19 pandemic?]. In the late 1970s, the American bioethicists Tom Beauchamp and James Childress described the four ethical principles that should guide a physician's actions in individual patient care. These principles are: (a) respect for autonomy; (b) doing well (beneficence); (c) not harming (non-maleficence); and (d) justice. In many countries, the global outbreak of SARS-CoV-2 has led to overloaded healthcare systems due to large numbers of COVID-19 patients. In order to provide care to this high volume of patients, far-reaching measures are taken that affect everyone. These measures are not taken from an individual patient's perspective but in the interest of public health; nonetheless, they can directly affect the individual patient's interests. This article examines the extent to which Beauchamp and Childress' ethical principles may be compromised during the COVID-19 pandemic.} \\
\textbf{\textcolor{orng}{Bad question:}} \textit{What are Beauchamp and Childress' ethical principles?}  \\ 
\textbf{\textcolor{rd}{Good question:}}  \textit{How does Beauchamp and Childress' ethical principles apply to the COVID-19 pandemic?}  \\
\midrule
\textbf{\textcolor{blu}{Document:}} {\footnotesize Covid-19 Confinement and Changes of Adolescent's Dietary Trends in Italy, Spain, Chile, Colombia and Brazil. Confinement due to the COVID-19 pandemic can influence dietary profiles, especially those of adolescents, who are highly susceptible to acquiring bad eating habits. Adolescents' poor dietary habits increase their subsequent risk of degenerative diseases such as obesity, diabetes, cardiovascular pathologies, etc. Our aim was to study nutritional modifications during COVID-19 confinement in adolescents aged 10 to 19 years, compare them with their usual diet and dietary guidelines, and identify variables that may have influenced changes. Data were collected by an anonymous online questionnaire on food intake among 820 adolescents from Spain, Italy, Brazil, Colombia, and Chile. The results show that COVID-19 confinement did influence their dietary habits. In particular, we recorded modified consumption of fried food, sweet food, legumes, vegetables, and fruits. Moreover, gender, family members at home, watching TV during mealtime, country of residence, and maternal education were diversely correlated with adequate nutrition during COVID-19 confinement. Understanding the adolescents' nutrition behavior during COVID-19 lockdown will help public health authorities reshape future policies on their nutritional recommendations, in preparation for future pandemics.} \\
\textbf{\textcolor{orng}{Bad question:}} \textit{What is the COVID-19 pandemic?}  \\ 
\textbf{\textcolor{rd}{Good question:}}  \textit{What is COVID-19 and how does it affect adolescents?}  \\
\midrule
\textbf{\textcolor{blu}{Document:}} {\footnotesize 67 CF patients with a declining FEV(1): At risk for acquisition of Burkholderia cepacia complex infection?. INTRODUCTION: Burkholderi
a cepacia complex (Bcc) infection is considered to be associated with worsening of CF lung disease. Patient to patient spread has been reported, however mechanisms of ac
quisition of Bcc are not well understood. Method: Data from the Belgian CF Registry (year 2000\u20132010) were collected. Inclusions: Bcc infected patients with entries
on lung function in at least 1 y before and 3 y after Bcc acquisition. For each case, we included 2 controls, matched for age at the index year (year of first Bcc infect
ion), pancreatic status, sex. Cumulative data up to 2 years before index year were compared to values obtained after infection using Rank sum test. Rate of decline in lu
ng function was adjusted for baseline lung function, age, sex. RESULTS: Bcc prevalence in CF is low in Belgium (<3\%). 183 patients were included: 61 cases, 122 controls.
 59\% were F508del homozygous. Mean age in cases was 20.9 y (SD 10.5) vs 20.3 y (SD 10.3) in controls. Among the Bcc, 54\% were unspecified, 31\% were B. multivorans. Mean
FEV(1) at index year was 65.2\% (SD 24.9) in cases vs 73.1 (SD 26.9) in controls (p = 0.07). FEV(1) decline before index year was significantly higher in cases (\u20131.7\%, SD 0.5) compared to controls (\u20131.0\%, SD 0.3) (p = 0.002). FEV(1) slopes were comparable in the period after index year (\u20131.1\%, SD 0.5, in cases vs \u20130.99\%, SD 0.4, p = 0.24). CONCLUSION: Our results suggest that a declining FEV(1) precedes acquisition of Bcc and may be a risk factor. After acquisition, lung function decline was comparable in Bcc infected and uninfected patients. These results should be interpreted with caution, since registry data are collected retrospectively and bear a risk of incompleteness or inaccuracy.} \\
\textbf{\textcolor{orng}{Bad question:}} \textit{What is the prevalence of Bcc in CF patients?}  \\ 
\textbf{\textcolor{rd}{Good question:}}  \textit{What is Burkholderia cepacia complex (Bcc) and how does it affect CF patients?}  \\
\midrule
\textbf{\textcolor{blu}{Document:}} {\footnotesize Safety and effectiveness of 5-mm and 10-mm electrothermal bipolar vessel sealers (LigaSure) in laparoscopic resections for sigmoid colon and rectal cancers.. BACKGROUND LigaSure\u2122 (Covidien, Mansfield, MA) has been used in cases undergoing laparoscopic colon and rectal resections. This study aims to analyze the efficacy and safety of the 5-mm and 10-mm devices. SUBJECTS AND METHODS Patients who received a laparoscopic or hand-assisted laparoscopic operation for a tumor located in the sigmoid colon or rectum since 2006 were abstracted from a prospectively designed database, and findings were analyzed in two groups based on size of the device used during the procedure. The videotapes of the procedures were watched, and operation reports were read to obtain further information on specific intra- and postoperative complications. Demographics, tumor and operation-related information, and postoperative data were compared. RESULTS Among 215 (128 [59.5\%] males; median age, 59.5\u00b113.8 years) patients, data obtained from the 5-mm (n=32) and 10-mm (n=183) groups were identical regarding demographics and data related to tumor (localization and stage) and operation (number of harvested lymph nodes, conversion rates, operation time, intraoperative bleeding, transfusion requirement, reoperation rates, complications, 30-day mortality, and length of hospital stay). However, more patients underwent an anterior resection in the 10-mm group than in the 5-mm group (31.7\% versus 15.6\%, P<.05). Further analyses found device-related bleeding in 8 (3.7\%) cases (2 [6.3\%] versus 6 [3.3\%] in the 5-mm versus 10-mm group, respectively, P>.05), requiring further attempts for hemorrhage control (n=6), conversion to open surgery (n=1), or relaparotomy (n=1). CONCLUSIONS The 5-mm and 10-mm LigaSure devices are similarly effective and safe during laparoscopic sigmoid colon and rectal resections. Severe bleeding from larger vessels may be observed, requiring conversion to open surgery or relaparotomy.", "question": "What is the safety and effectiveness of the LigaSure device?} \\
\textbf{\textcolor{orng}{Bad question:}} \textit{What is the LigaSure device?}  \\ 
\textbf{\textcolor{rd}{Good question:}}  \textit{What is the safety and effectiveness of the LigaSure device?}  \\
\bottomrule
\end{tabularx}
 \end{small}
\end{table*}

\begin{table*}[htbp!]
 \begin{small}
\setlength\tabcolsep{6pt}
 \caption{Examples of questions generated from the Natural Questions dataset using the Curie model with generated good and bad questions.}
 \label{tab:examples_nq}
\begin{tabularx}{\textwidth}{X}
\toprule
\textbf{\textcolor{blu}{Document:}} {\footnotesize 1993 Miami Dolphins season. The season was marked by Don Shula passing George Halas's record for most wins, against the Philadelphia Eagles. Also, during the Week 5 game against Cleveland, quarterback Dan Marino ruptured his Achilles' tendon and was lost for the remainder of the season. Quarterback Scott Mitchell filled in for Marino, and was Player of the Month for October 1993. Mitchell, too, became injured, leaving the then 9\u00e2\u20ac\u201c2 team in the hands of Doug Pederson and NFL veteran Steve DeBerg.} \\
\textbf{\textcolor{orng}{Bad question:}} \textit{What happened to Dan Marino?}  \\ 
\textbf{\textcolor{rd}{Good question:}}  \textit{What happened to the 1993 Miami Dolphins?}  \\
\midrule
\textbf{\textcolor{blu}{Document:}} {\footnotesize Social construction of gender. On Butler's hypothesis, the socially constructed aspect of gender performativity is perhaps most obvious in drag performance, which offers a rudimentary understanding of gender binaries in its emphasis on gender performance. Butler understands drag cannot be regarded as an example of subjective or singular identity, where \"there is a 'one' who is prior to gender, a one who goes to the wardrobe of gender decides with deliberation which gender it will be today\".[34]:21 Consequently, drag should not be considered the honest expression of its performer's intent. Rather, Butler suggests that what is performed \"can only be understood through reference to what is barred from the signifier within the domain of corporeal legibility\".[34]:24} \\
\textbf{\textcolor{orng}{Bad question:}} \textit{What is drag?}  \\ 
\textbf{\textcolor{rd}{Good question:}}  \textit{What is the social construction of gender?}  \\
\midrule
\textbf{\textcolor{blu}{Document:}} {\footnotesize Wentworth Miller. In 1995, Miller relocated to Los Angeles to pursue an acting career.[10] He has stated that his rocky road to stardom \"was a long time in the coming and there were a lot of upsets and a lot of failures and roadblocks, but I couldn't walk away from it. I needed it like I needed air, it was just something I had to do.\"[11] Miller's first TV appearance was as student-turned-sea monster Gage Petronzi on Buffy the Vampire Slayer (\"Go Fish\", 1998).} \\
\textbf{\textcolor{orng}{Bad question:}} \textit{What is Wentworth Miller's first TV appearance?}  \\ 
\textbf{\textcolor{rd}{Good question:}}  \textit{Information on Wentworth Miller's acting career.}  \\
\midrule
\textbf{\textcolor{blu}{Document:}} {\footnotesize Sport in Argentina. The national team has won the FIFA World Cup twice (in 1978 and 1986), successive Olympic gold medals (in 2004 and 2008), fourteen Copas Am\u00e9rica, one Confederations Cup and six World Youth Championships. Argentine clubs have won the Copa Libertadores, the top continental competition, a record 22 times and the Intercontinental Cup or FIFA Club World Cup 8 times, a record shared with Brazilian clubs. The Argentine Primera Divisi\u00f3n is the top level domestic competition. The country's most famed football idol is Diego Maradona. Actually the most famous football player is Lionel Messi.", "question": "Information on the Argentine national team and their history.} \\
\textbf{\textcolor{orng}{Bad question:}} \textit{What is the most famous football player in Argentina?}  \\ 
\textbf{\textcolor{rd}{Good question:}}  \textit{Information on the Argentine national team and their history.}  \\
\midrule
\textbf{\textcolor{blu}{Document:}} {\footnotesize American game show winnings records. In 1999, McKee was passed by Michael Shutterly, who was the biggest winner in the first airing of Who Wants to Be a Millionaire in the United States. Shutterly was the first contestant on the show to get to the 15th and final question, but elected to walk instead with \$500,000 (equivalent to \$719,000 in 2016),[1] which made him the biggest winner in American game show history at the time. Shutterly had previously won \$49,200 as a 4 day champion on Jeopardy! in 1988, making his career winnings total \$549,200. [12]} \\
\textbf{\textcolor{orng}{Bad question:}} \textit{What is the biggest winner in the history of American game shows?}  \\ 
\textbf{\textcolor{rd}{Good question:}}  \textit{Who is the biggest winner in the history of American game shows?}  \\
\midrule
\textbf{\textcolor{blu}{Document:}} {\footnotesize New Zealand Warriors. The Warriors reached their zenith to date in the National Rugby League season 2002. They won the Minor Premier
ship, finishing in first place at the conclusion of the regular season after the Bulldogs lost 37 competition points late in the season due to severe salary cap breaches
. The club played what stands as the first finals match to have been held outside Australia at Mt Smart Stadium in the first week of the Finals Series. The Warriors woul
d defeat their bogey side Canberra 36\u00e2\u20ac\u201c20 after surviving an early scare.} \\
\textbf{\textcolor{orng}{Bad question:}} \textit{What is the National Rugby League?}  \\ 
\textbf{\textcolor{rd}{Good question:}}  \textit{Information on the New Zealand Warriors.}  \\
\bottomrule
\end{tabularx}
 \end{small}
\end{table*}

\end{document}